\documentclass[letterpaper, 10 pt, journal, twoside]{IEEEtran}  

\IEEEoverridecommandlockouts                              

\newcommand{\subparagraph}{}
\usepackage{titlesec}

\usepackage{graphicx}
\usepackage[font=footnotesize]{caption}
\usepackage{subcaption}
\usepackage{titlesec}
\usepackage{color}
\usepackage{cite}   

\usepackage{longtable}
\usepackage{verbatim}
\usepackage{array}
\usepackage{ragged2e}
\newcolumntype{C}[1]{>{\centering\arraybackslash}b{#1}}

\title{
Object-Independent Human-to-Robot Handovers using Real Time Robotic Vision
}
\author{Patrick Rosenberger$^{1,5}$,
        Akansel Cosgun$^{2,5}$,
        Rhys Newbury$^{2,5}$,
        Jun Kwan$^{2,5}$,
        Valerio Ortenzi$^{3}$, \\
        Peter Corke$^{4,5}$ and
        Manfred Grafinger$^{1}$
\thanks{Manuscript received: June, 1, 2020; Revised August, 15, 2020; Accepted September, 7, 2020.}
\thanks{This paper was recommended for publication by Hong Liu upon evaluation of the Associate Editor and Reviewers' comments.}
\thanks{$^{1}$ TU Wien, Austria {\tt\small patrick.m.rosenberger@gmail.com}}
\thanks{$^{2}$ Monash University, Australia}
\thanks{$^{3}$ University of Birmingham, United Kingdom}
\thanks{$^{4}$ Queensland University of Technology, Australia}
\thanks{$^{5}$ Australian Center of Robotic Vision, Australia}
\thanks{$^{A}$ \texttt{https://patrosAT.github.io/h2r\_handovers/}}
\thanks{Digital Object Identifier (DOI): see top of this page.}
}

\begin{document}

\maketitle


\begin{abstract}


We present an approach for safe and object-independent human-to-robot handovers using real time robotic vision and manipulation. We aim for general applicability with a generic object detector, a fast grasp selection algorithm and by using a single gripper-mounted RGB-D camera, hence not relying on external sensors. The robot is controlled via visual servoing towards the object of interest. Putting a high emphasis on safety, we use two perception modules: human body part segmentation and hand/finger segmentation. Pixels that are deemed to belong to the human are filtered out from candidate grasp poses, hence ensuring that the robot safely picks the object without colliding with the human partner. The grasp selection and perception modules run concurrently in real-time, which allows monitoring of the progress. In experiments with 13 objects, the robot was able to successfully take the object from the human in 81.9$\%$ of the trials.

\end{abstract}

\begin{IEEEkeywords} 
Physical Human-Robot Interaction,
Perception for Grasping and Manipulation,
Deep Learning in Grasping and Manipulation
\end{IEEEkeywords} 

\vspace{-0.1cm}
\section{Introduction} 
\label{section_introduction}
\IEEEPARstart{P}{assing}
objects is a collaborative manipulation task that comes naturally to people. In contrast, it is challenging for robotic systems to execute handovers as safely and fluently as humans. This difference can be attributed to how differently humans and robots perceive their environment, deal with uncertainty, and react to previously unseen situations~\cite{ortenzi2019robotic}. A smooth interaction requires the interaction partners to coordinate their movements and to react to changes in intent and positioning in real time, regardless of the environment and the object that is passed. 

The topic of object handovers between humans and robots is an active area of research thanks to the interest in collaborative robots in the recent years \cite{ajoudani2018progress}. There are two important attributes for successful handovers: safety and generalizability. For safety, it is not advisable that the robot comes into contact with a human as this may lead to injuries. Furthermore, negative experiences or subjective perceptions limit a human's willingness to cooperate with robots, thus impeding collaboration \cite{salem2015would}. In the past, safety was realised by physically separating the human's and the robot's spheres of action \cite{kruger2009cooperation}, which is not possible for handovers, as the workspaces of the interaction partners have to intersect for the object exchange. The second attribute is generalizability. For the handovers to be useful, the robot should be capable of exchanging any object that it is physically able to grasp. This requires the robot to adapt its motions even for objects it has not seen before. 

The direction of object handovers can be from robot to human and from human to robot. In the case of robot-to-human handovers, the robot extends the object to the human, who picks up the object from the robot’s end effector. Therefore, it is mostly the human's responsibility to differentiate between the object and the interaction partner, decide how to grasp the object and adapt to changes in how the object is presented. The roles are reversed for human-to-robot handovers as the robot has the responsibility of grasping the object from the human while ensuring safety.


\begin{figure}[t!]
    \centering
     \includegraphics[trim={1.5cm 5.7cm 2cm 1.9cm}, clip, width=1.0\columnwidth]{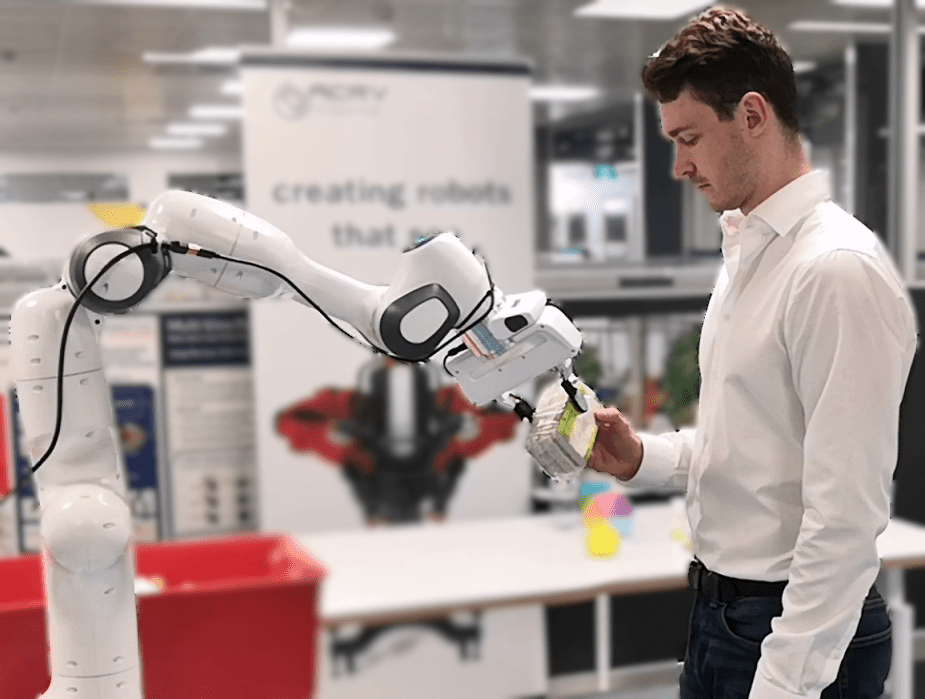}
    \caption{Our approach is capable of grasping various objects from humans.}
    \vspace{-0.3cm}
    \label{fig:intro}
    \vspace{-0.3cm}
\end{figure}

This paper presents an approach for object-independent human-to-robot handovers. Our system is capable of grasping previously unseen objects from the human partner's hand while avoiding collisions by coupling real-time perception with real-time manipulation. As the robot is approaching the object, our grasping approach continuously monitors its interaction partner to prevent the grabbing or pinching of the human body, hand and fingers. The contributions of our work are as follows: 
\begin{itemize}
\setlength\itemsep{0pt}
    \item An object-independent approach for human-to-robot handovers using real-time grasp pose selections.
    \item Use of real-time hand/finger detection so that the robot avoids grasping any human part during the interaction.
    \item Evaluation of the approach with 13 objects and 520 total handover trials, and further testing of edge cases with more objects
    \item Public release of the code to the community.$^{A}$
\end{itemize}

The paper is organized as follows. After reviewing the literature in Sec.~\ref{section_related_work}, we outline our design principles in Sec.~\ref{section_principles}. The hardware setup is described in Sec.~\ref{section_hardware}, followed by our approach in Sec.~\ref{section_handover}. The experiments and results are presented in Sec.~\ref{section_evaluation}, before concluding in Sec.~\ref{section_conclusion}.

\section{Related Work} 
\label{section_related_work}

Human-to-robot handovers have become an important topic of research within human-robot interaction. In this section we focus on previous approaches that cover the complete human-to-robot handover pipeline. In contrast, \cite{ortenzi2020object} provides an in-depth review of the individual topics related to object handovers in robotics.

Previous work in the literature has specialised on specific aspects of the exchange in both the social-cognitive and the physical layers of the interaction \cite{strabala2013toward}.
For example, Strabala et al. \cite{strabala2013toward} study how humans pass objects among themselves and how the coordination process can be transferred to robot-to-human and human-to-robot handovers. The authors implemented a hard-coded robotic system that hands an object to a human user or takes it from them. 
Medina et al. \cite{medina2016human} develop a handover controller that models human-like dynamics and is capable of estimating the most likely point of handover based on the human behavior. The authors test the approach using a robot arm with a 5-finger gripper that measures the interaction force during the handover of a plastic bottle. The system requires markers that are attached to the human partner's wristband.
Meyer zu Borgsen et al. \cite{meyer2017hand} implement a wrist-force based handover detection approach to investigate the impact of prior knowledge on the handover interaction. The authors show that trained users have a significantly higher handover success rate as they know how to interact with the robot to increase the likelihood of a successful interaction. Bianchi et al. \cite{bianchi2018touch} present a touch-based approach for the vision-independent autonomous grasping of objects during human-to-robot handovers. Upon recognizing the touch of an object, the soft hand starts to slowly move towards the object, slowly closing the gripper. Thereby, the optimal grasp pose is calculated and constantly updated using the touch-based information of the object's surface. 
Vogt et al. \cite{vogt2018one} introduce an imitation methodology that allows robots to learn interaction patterns based on observed interactions between two humans. The feasibility of the approach is evaluated while passing different large cuboid objects.
Pan et al. \cite{pan2018evaluating} investigate the social perception of robots with respect to human-to-robot handovers. An electromagnetic gripper is used that grabs a handover baton with markers attached. Based on these results, Pan et al. \cite{pan2019fast} study the effects of speed and reaction time on the perceived handover quality. The robot system uses a soft hand to grab a ring with a diameter of 30cm. The precise matching is done with the help of markers. 
Nemlekar et al. \cite{nemlekar2019object} conduct an extensive motion study to predict the likely object transfer point based on the current human behavior. To evaluate this method, the authors develop a robot system that hands over a fully defined object.

This review shows that the focus of previous research on human-to-robot handovers has been primarily on individual aspects of the exchange. In contrast, our work provides a full pipeline for human-to-robot handovers. Our work shows some similarities to \cite{yang2020human}. However, the adopted perception and grasp selection approaches are completely different. Yang et al.\cite{yang2020human} classifies human grasp poses and adapts the robot's trajectory accordingly, \textit{i.e.}, their neural network can detect seven different grasp poses and grab a held cube at its centre. Alternatively, our approach implements an object-independent grasp planning algorithm that selects the best picking location based on the object's shape and therefore allows the transfer of unknown objects. Furthermore, our approach tracks the human body and the human hand to avoid unwanted contact with the human partner (thus ensuring safety during the exchange of the object). Humans seem to show a high degree of adaptation and usually help the robot to grasp the object \cite{Edsinger2007} like with offering it in a configuration favourable for grasping \cite{chan2015characterization}. However, we believe that our method improves the handover performance by decreasing the possibility of an unwanted contact with the hand/body of the human passer, thus increasing the safety of the exchange.

\section{Design Principles} 
\label{section_principles}

\textbf{Safety:} Safety is critical for any human-to-robot interaction scenario. While incidental contact may occur during handovers, especially for small objects, only low-impact contact between the human and robot is allowed. Moreover, the robot should not pinch the human's hands with its gripper. Given these conditions, the system must have the functionality to differentiate the human from the object. In our work, human hands and body parts are detected to minimize the risk of collisions and pinching.

\textbf{Object independence:} The usefulness of handovers is directly related to the number and type of objects that can be passed. A system that can only handle a small set of known objects can only be used in controlled environments. A general-purpose system requires the robot to accept various previously unseen objects. In our work, we aim for object generalization by using a generic object detector and a learning-based grasping pose selection approach trained on everyday objects.

\textbf{Object presentation:} Human-human handover studies show that humans present objects in a variety of ways depending on the object affordances~\cite{cini2019choice}. The usefulness of human-to-robot handovers increases if the robot can take objects presented in various ways, such as on the palm, handed over with both hands, or when the object or the hand is not stationary. With our approach, the robot is able to handle various presentation modes, as shown in Fig~\ref{fig:holds}.

\textbf{Environment:} The need for artificial adaptions of the environment, such as markers and external cameras, negatively impacts the general applicability of robotic systems. Therefore, such systems should contain the required hardware on-board. Our system utilizes a single, end-effector-mounted RGB-D camera and does not rely on artificial markers.

\textbf{Real-time interactions:} A study by Pan et al. \cite{pan2019fast} shows that real-time motions with deliberate delays increase the natural feeling of the handover, thereby increasing the human partner's comfort. Hence, the interaction speed is of critical importance for the fluidity of the interaction as the robotic system should be reactive to the changes in the position of the human hand as well as of the object. This necessitates that the perception and control modules operate in real-time. In our work, the computer vision and grasping modules process the incoming images in real-time.

\section{System Setup} 
\label{section_hardware}

For this work a desk-mounted Franka-Emika Panda robotic arm is used. The robot has 7 degrees of freedom, a maximum reach of 855mm, and a two-finger parallel gripper as its end effector. To increase the grasp robustness, we used custom-made gripper jaw surfaces made of silicon rubber that are based on \cite{guo2017design}. Fully open, the gripper can pick objects with a maximal width of 7cm. The arm's maximum payload is 2.3kg with the end effector's weight accounted. An Intel RealSense D435 RGB-D camera, with resolution 640x480 at 30fps, is mounted to the robot's end effector. The camera has a blind spot for distances closer than 0.105m. The system is implemented on a distributed network consisting of five computers that are equipped with Nvidia GPUs with at least 8GB memory, running on 64-bit Ubuntu and connected via a LAN. ROS is used for inter-node messaging and interfacing with the robot and the camera. The vision and the grasping modules use computers with dedicated GPUs in order to satisfy the real-time requirements. One of the computers is used to host the control module for the Panda robot.

\section{Approach} 
\label{section_handover}

\begin{figure}[h!]
    \centering
    \vspace{-2pt}
    \includegraphics[width=1.0\columnwidth]{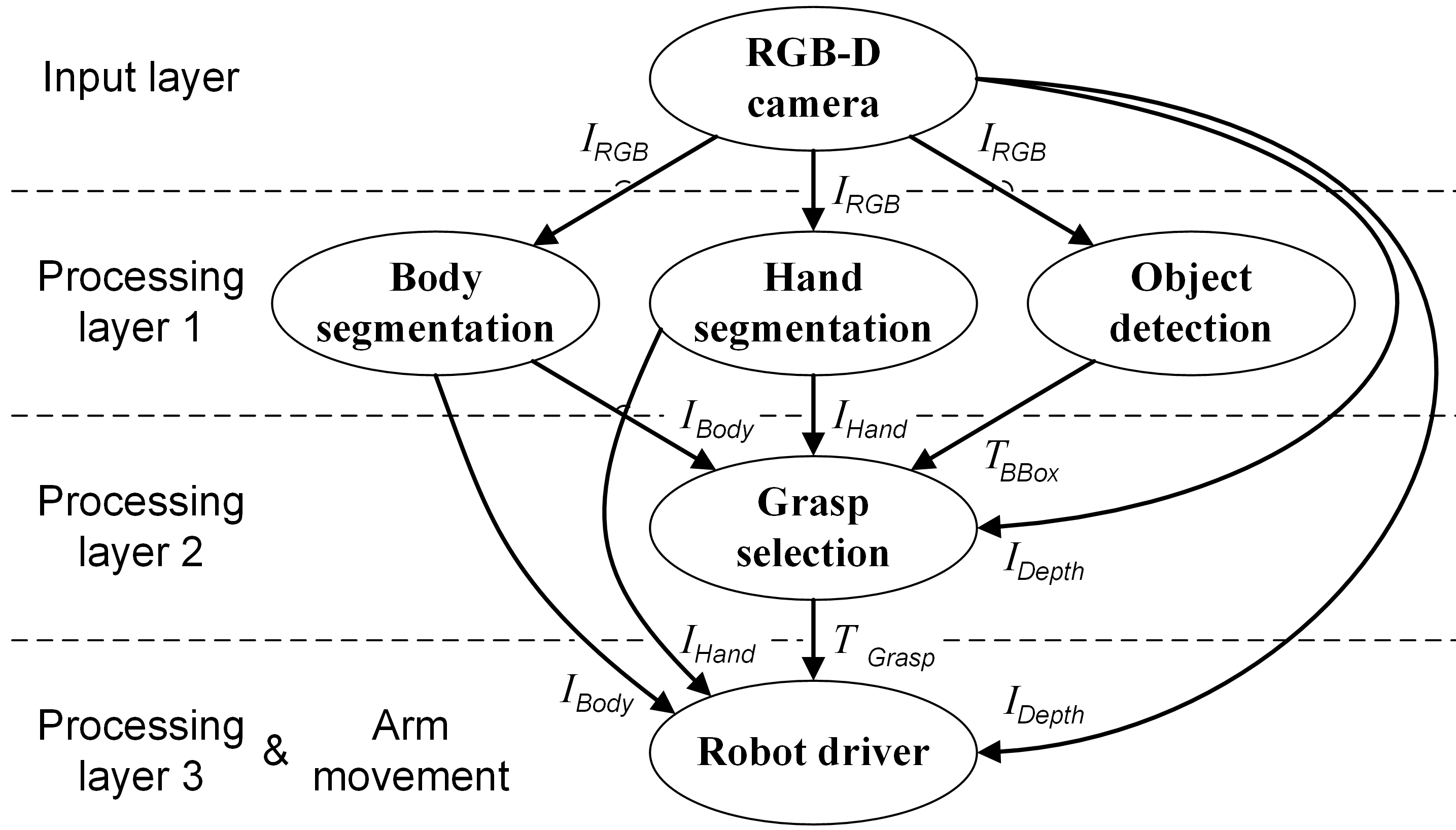}
    \caption{System Diagram
        (\textit{I\textsubscript{RGB}}: RGB image; 
        \textit{I\textsubscript{Depth}}: depth image; 
        \textit{T\textsubscript{BBox}}: pixel coordinates of the objects' bounding boxes (tensor);
        \textit{I\textsubscript{Body}}: segmented mask of the body;
        \textit{I\textsubscript{Hand}}: segmented mask of the hand;
        \textit{T\textsubscript{Grasp}}: coordinates of the best grasp point, gripper orientation, and expected success rate (tensor)).}
    \label{fig:approach}
    \vspace{-4pt}
\end{figure}

Our approach spans six modules that are structured into one input and three processing layers (see Fig.~\ref{fig:approach}). Modules of the same layer operate in parallel, while modules of different layers in sequence; the arrows illustrate the data flow between the modules. Each module operates with a frame rate between 24 and 30 fps, the whole pipeline requires 0.0625 seconds to process each image. To reduce the time between image capturing and arm movement, all modules are event driven and immediately process an input once it becomes available. This allows earlier modules to process new input while later modules are still processing the former one. Buffers between the modules make the system more robust.
The grasp point is calculated over a window of five frames, leading to a minimal handover initiation time of 0,19563 seconds. This value is comparable to the normal reaction times of people in traffic \cite{schmidtke2013untersuchungen}. Frames in which the grasp point deviates more than 7cm from the mean in any direction are discarded. In case too many frames are discarded, a new window is selected. This ensures safe handovers, but deteriorates the handover initiation time in case of too much noise. There is no time limitation on the handover initiation speed. Rather, the handover is considered a failure once the human interaction partner becomes impatient and aborts the handover. For safety considerations, the robot arm movement slows down significantly during the last 7cm of the handover. This allows the human interaction partner to react if s/he feels unsafe.

The RGB-D camera mounted at the end effector provides the images that are used by all of the perception modules which run concurrently in real-time (\textit{I\textsubscript{RGB}: RGB image, \textit{I\textsubscript{Depth}}: depth image)}. The robot starts the handover at a fixed home pose, which is positioned so that the human interaction partner and the object are within the field of view of the camera. First, the object detector (Sec.~\ref{sec:object_detector}) detects the objects in the RGB image and outputs the corresponding bounding boxes (\textit{T\textsubscript{BBox}}). We deliberately select a low detection threshold so that the module acts as a generic detector. The foremost object within the robot's reach is chosen as the handover target. We insert an imaginary plane to the depth image right behind the object in order to mimic tabletop grasping, hence facilitating the selection of the grasp point. The grasp selection module (Sec.~\ref{sec:grasp_selection}) outputs a grasp quality estimation along with the associated grasp orientation and gripper width for each pixel in the depth image (\textit{T\textsubscript{Grasp}}). The hand segmentation (Sec.~\ref{sec:hand_segmentation}) and body segmentation (Sec.~\ref{sec:body_segmentation}) modules each output a pixel-wise mask (\textit{I\textsubscript{Hand}}, \textit{I\textsubscript{Body}}), which are used to filter out any potential grasp points that belong to the user's body, hand, or fingers.
As both algorithms do not offer 100\% accuracy, using two neural networks with different architectures that are trained on different datasets concurrently introduces redundancy, increases the system's robustness and decreases the chance of misclassifications.
Finally, the grasp point with the highest estimated success likelihood is chosen and translated into the robot's base frame. The robot driver module (Sec.~\ref{sec:robot_control}) moves the end effector towards the selected grasp point via visual servoing. The segmentation masks are updated in real-time to dynamically handle the changes in the hand/body positions. Once the object handover is complete, the robot arm is moved over a bin and the object is dropped. We assume that a cooperative human partner is available for the handover.

\subsection{Object Detection}
\label{sec:object_detector}

\begin{figure}[t!]
    \begin{subfigure}{0.49\columnwidth}
        \includegraphics[width=\textwidth]{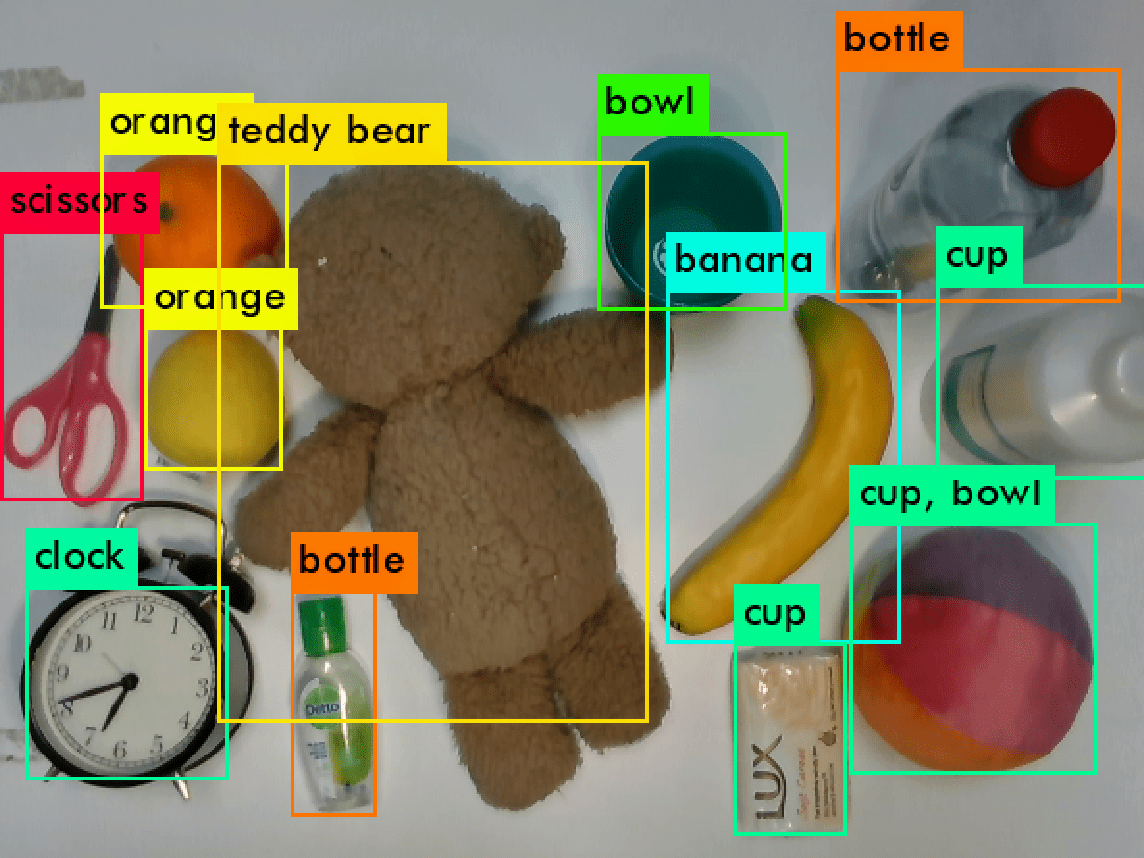}
        \caption{}
        \label{fig:yoloTable}
    \end{subfigure} 
    \begin{subfigure}{0.49\columnwidth} 
        \includegraphics[width=\textwidth]{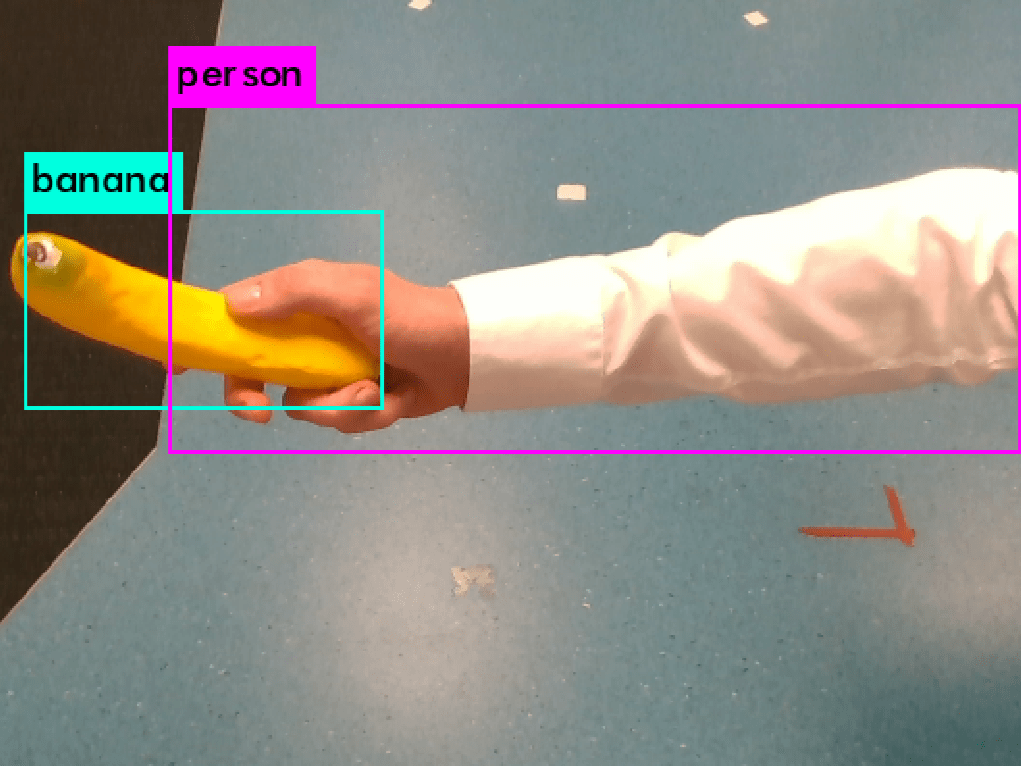}
        \caption{}
        \label{fig:yoloHand}
    \end{subfigure}  
    \vspace{-4pt}
    \caption{
        a) YOLO object detector used with several objects on a table. b) YOLO object detector used with an object in the user's hand.
    }
    \label{fig:yolo}
    \vspace{-8pt}
\end{figure}

This module is used for finding the pixel coordinates of the object(s) that the human partner is holding, if any (see Fig.~\ref{fig:yolo}). It accepts an RGB image and outputs the bounding boxes of all the objects found in the image. We use a third-party implementation of the YOLO v3 object detector~\cite{redmon2018yolov3}, trained on the COCO dataset~\cite{lin2014microsoft} consisting of 80 different object categories. Since our goal is to enable handovers for any class of objects, we select a detection confidence threshold low enough to deliberately enable misclassifications for objects that do not belong to one of the 80 categories, but high enough to prevent false detections of non-existent objects. From the detected objects, we filter out the ones outside the robot's workspace. This is done by calculating the mean distance of the depth image pixels for each object bounding box. We further remove any object with the label ``Person". To prevent any unintended movement, the robot does not start moving until a graspable object is detected within the robot's workspace. Once the robot starts moving towards the object, the output of this module is no longer used.

\subsection{Body Segmentation}
\label{sec:body_segmentation}

This module outputs a per-pixel segmentation that corresponds to the individual body parts (head, torso, upper arm, forearm including hand, thigh, and lower leg including foot) of all people in the image, given an RGB image as input. We treat all body parts equally, as any contact between the human and robot should be avoided if possible. Fig.~\ref{fig:bodyFront} shows a regular picking scene from the robot's perspective with all pixels not detected as part of the human body blacked out. We use a third-party implementation of a lightweight version of RefineNet originally proposed by Lin \cite{lin2017refinenet}, which was trained on the portion of the PASCAL dataset that belong to the class ``Person". As this network detects body parts from a holistic perspective, it is not specialized in detecting fingers under occlusion. To compensate for this limitation, we implement a separate hand segmentation module, described in the next subsection. 

\subsection{Hand Segmentation}
\label{sec:hand_segmentation}

\begin{figure}[t!]
    \begin{subfigure}[]{0.49\columnwidth}
        \includegraphics[width=\textwidth]{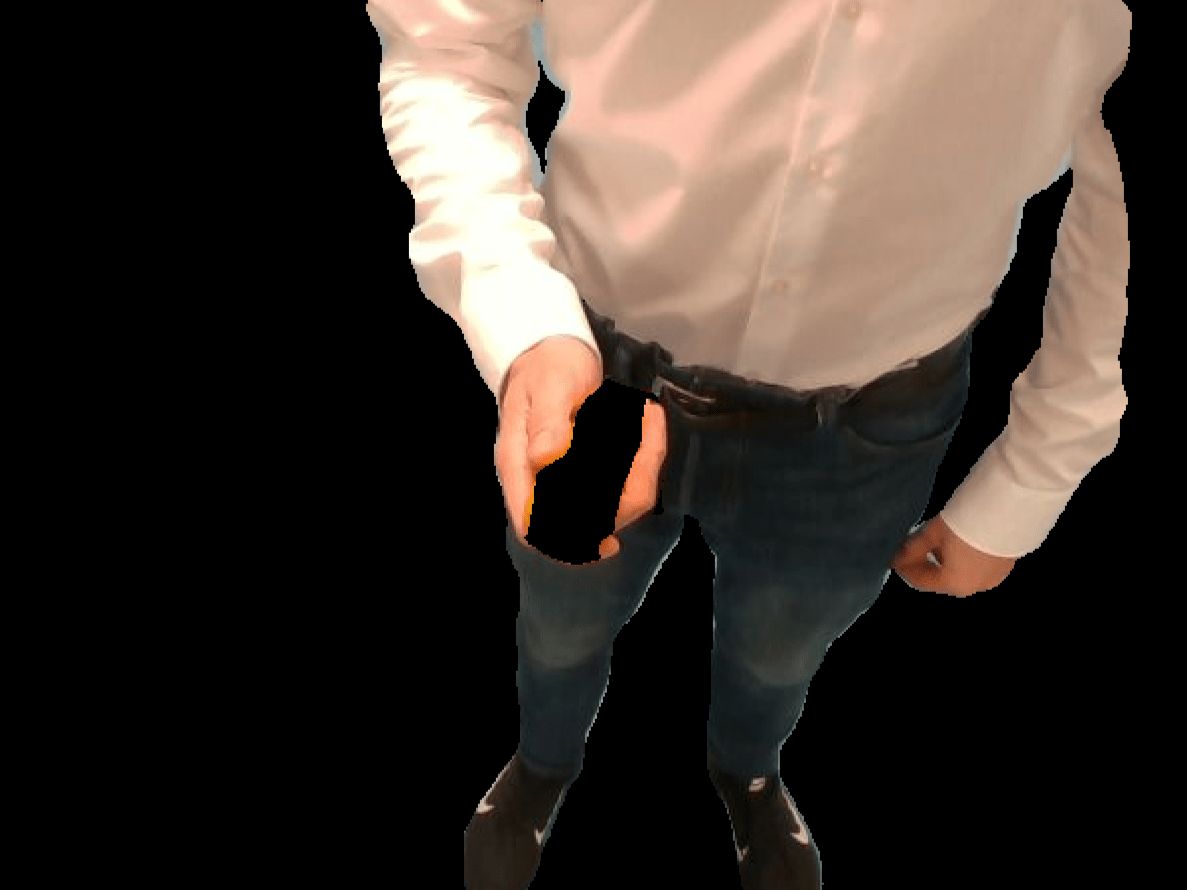}
        \caption{}
        \label{fig:bodyFront}
    \end{subfigure} 
    \begin{subfigure}[]{0.49\columnwidth} 
        \includegraphics[width=\textwidth]{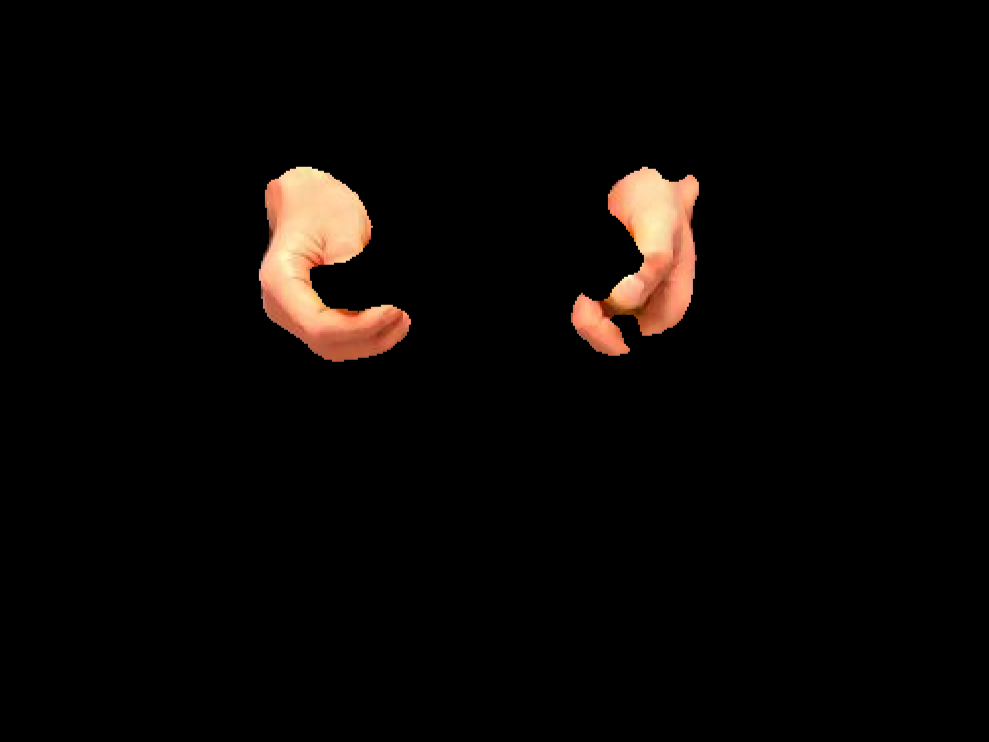}
        \caption{}
        \label{fig:handFront}
    \end{subfigure}  
    \vspace{-4pt}
    \caption{
        Segmentation results from the eye-in-hand perspective with pixels not belonging to the corresponding classes blacked out. a) Body segmentation showcasing the detection of all body part pixels with the user holding an object firmly in one hand. b) Hand segmentation showcasing the detection of only hand pixels with the user holding an object in two hands.
    }
    \label{fig:body_hand}
\end{figure}

This module takes the RGB image as its input and outputs a per-pixel segmentation with two classes: hand and not-a-hand. The goal for this module is to detect any pixels belonging to the hand of the interaction partner and in particular fingers, even when only parts such as a single finger tip are visible. The information from this module is used to avoid grasping the human's hand or fingers. Fig.~\ref{fig:handFront} shows a picking scene from the robot's perspective with all pixels not belonging to the human hand blacked out. We implemented a pyramid scene parsing network originally proposed by Zhou et al. \cite{zhou2017scene}: a ResNet with 50 layers and dilated network strategy is utilized along with a pyramid pooling module. The EgoHands dataset \cite{egohands} is used to train the network, which contains 48 videos with a total of 4800 segmented frames that are collected from a first-person view by a Google Glass device. In these videos, subjects are sitting across each other while they are engaging in several activities in different environments. 4400 images are used for training, the rest are used for validation. The network's performance is measured with two metrics: 1) mean Intersection-Over-Union (mIoU) which measures the degree of overlap between the ground truth and detected pixels, and 2) pixel accuracy, which is the percentage of pixels being classified correctly. The trained model achieved a mIoU of 0.897 and a pixel accuracy of 0.986 on the validation set.

The network is trained on a dataset consisting of hands only. During testing we observed that it is also susceptible to detect human skin. We decided to keep this unintended feature as detecting additional pixels belonging to human skin does not affect the system's performance and contributes to system safety.


\subsection{Grasp Selection}
\label{sec:grasp_selection}

\begin{figure}[t!]
    \begin{subfigure}{0.49\columnwidth}
        \includegraphics[width=\textwidth]{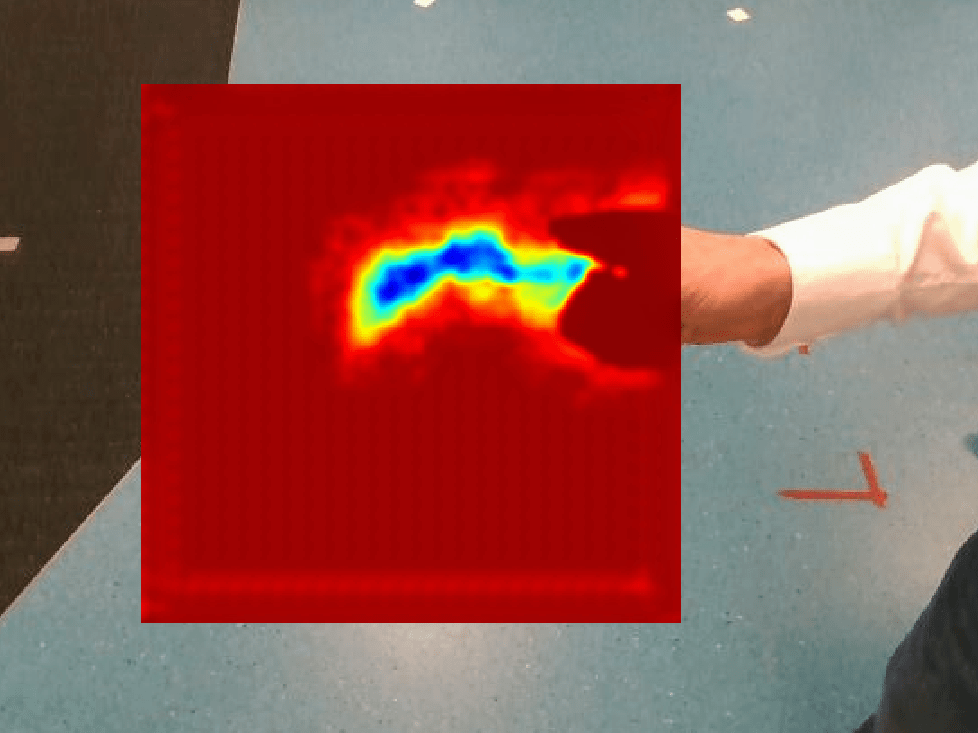}
        \caption{}
        \label{fig:ggcnn}
    \end{subfigure} 
    \begin{subfigure}{0.49\columnwidth} 
        \includegraphics[width=\textwidth]{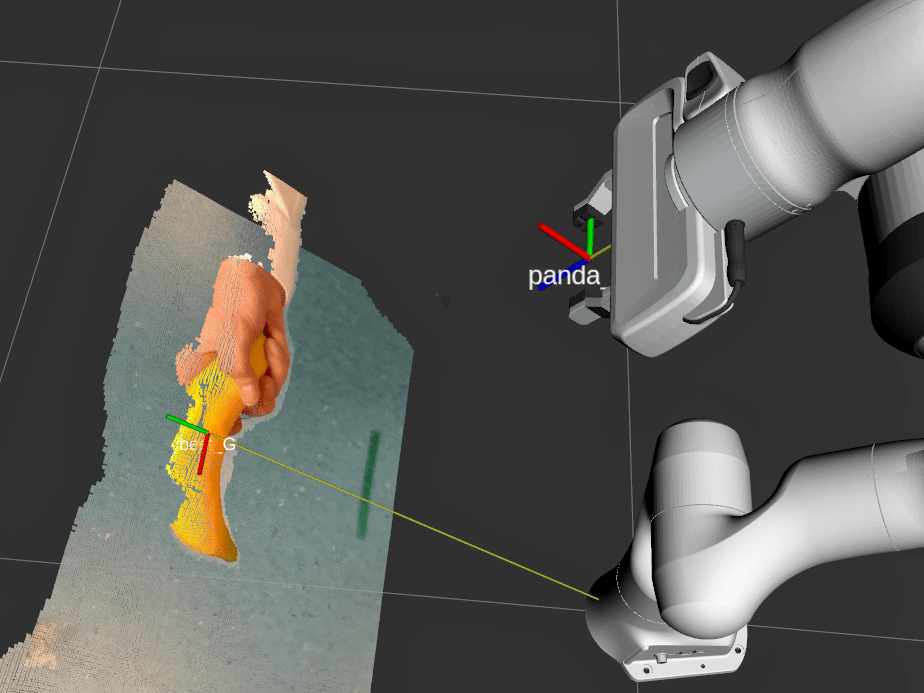}
        \caption{}
        \label{fig:graspPoint}
    \end{subfigure}  
    \vspace{-4pt}
    \caption{
        a) Heat map showing the expected grasp probabilities for each point (red - low probability, blue - high probability). The narrowed field of view is due to chopping operations required by the GG-CNN. b) Selected grasp point and orientation in the robot's base frame.
    }
    \label{fig:ggcnnGraspPoint}
\end{figure}

Our grasp selection module inputs the depth image, bounding boxes and segmented masks of the previous modules and calculates a map of possible grasps (see Fig~\ref{fig:ggcnn}). Among these, the module selects the grasp point with the highest expected success rate for a 2-finger gripper (see Fig~\ref{fig:graspPoint}) and outputs it with respect to the robot's base frame. The grasping approach is based on the Generative Grasping Convolutional Neural Network (GG-CNN)~\cite{morrison2018closing}. GG-CNN generates grasp poses on a pixelwise basis using an input depth image from an eye-in-hand camera and runs in real-time as it has orders of magnitude fewer parameters than other CNNs used for grasp synthesis. GG-CNN is trained on a dataset created from the Cornell Grasping Dataset~\cite{lenz2015deep}, which consists of singular objects laying on a table. We could not use the original GG-CNN for two reasons:

\begin{itemize}
\setlength\itemsep{0pt}
    \item It is trained on a background with a flat surface (top-down view), while in our application the objects are hand-held.
    \item It does not distinguish between human fingers and the object, hence it may suggest grasping the hand or fingers.
\end{itemize}

We therefore use GG-CNN with the following two modifications: 1) we pre-process the depth image by inserting an imaginary plane to the depth image right behind the target object and in front of the human interaction partner, essentially replacing the background with a planar surface. The plane's orientation is parallel to the camera's optical axis, preventing pixels that are behind the foremost object from being considered by the grasp selection algorithm. The depth image looks as if there is a single object laying on a table, making it look similar to the training set images. The modified depth image is then processed by the GG-CNN which outputs a proposed 2D grasp position and  orientation (assuming a top-down grasp from the camera's perspective) for each pixel, along with their predicted grasp qualities. 2) The pixels that belong to the human, as predicted by the body and hand segmentation modules, are filtered out to ensure a safe handover. Moreover, grasp points that are in close proximity to the human hand or any other body part (5 pixels) are also filtered out using a morphological operation erosion. 

Once the grasp point with the highest predicted success rate is selected, along with the associated grasp orientation and gripper width that GG-CNN has predicted, the grasp pose is transformed into the robot's base coordinate frame, before it is passed to the robot control module. 

\subsection{Robot Control}
\label{sec:robot_control}

This module translates the output of the grasp selection module into joint and gripper actions. The process starts by selecting the point of handover based on the optimal grasp points as determined by GG-CNN. The selection is done over a window of five frames, discarding each frame with a deviation of more than 7cm from the mean in any direction. Finally, the mean in x,y and z of the remaining frames is selected as point of handover. This approach prevents noise from having a negative impact. Once chosen, the grasp pose stays the same until the handover is completed. Due to the high computational load of the combined pre-processing-pipeline and the latency while working with several ROS nodes on different computers, updating the grasp point with each frame is not feasible. The handover starts once the grasp pose is selected. We use Position-Based Visual Servoing ~\cite{hutchinson1996tutorial} to drive the robot to the grasp pose and the velocity commands are sent to the low-level Panda robot controller. Once the robot reaches the grasp pose, it closes its gripper, moves over the dropping location and opens the gripper.

As the robot approaches the object, we monitor the grasp pose to check for potential errors. We compare the current distance to the object with the expected distance and abort the handover if the deviation exceeds a defined error margin or if a human body part moves too close to the grasp position. In such cases, the robot moves back to its initial position and the handover is restarted. Furthermore, if the low-level robot controller reports an error, or a collision is sensed by the joint forces, we execute a recovery behavior: the robot is stopped, the gripper is opened, and after waiting three seconds, the arm returns to the home pose. 

\section{Pilot Study} 
\label{section_evaluation}

We demonstrate the effectiveness of our approach with a set of experiments. We present the experiment setup in Sec.~\ref{sec:handover_experiments}, the quantitative results in Sec.~\ref{sec:evaluation_results} and qualitative results from further testing in Sec.~\ref{sec:limit_internal_test}.

\begin{table*}[ht!]
    \begin{tabular}{|l|C{1.05cm}|C{1.05cm}|C{1.05cm}|C{1.25cm}|C{1.05cm}|C{1.05cm}|C{1.05cm}|C{1.05cm}|C{1.05cm}|C{1.05cm}|C{1.05cm}|C{1.05cm}|C{1.05cm}|C{1.05cm}|C{1.05cm}|}
        \hline
                         & Spam & Banana & Lemon &    Strawberry & Peach & Pear & Plum &  Mustard & Tuna & Sugar &   Biscuits & Custard & Jelly & Overall   \\
        \hline
        Successful & 92.5 & 80 & 80 & 75 & 85 & 92.5 & 85 & 80 & 82.5 & 80 & 75 & 82.5 & 75 &81.9\\
        \hline
        Safety Stop & 7.5  & 10 & 7.5 & 10    & 2.5  & 5  & 10  & 5  & 0  & 0  & 0   & 7.5  & 10 & 5.8\\
        \hline
        Grasping Fail & 0  & 10 & 12.5 & 12.5    & 12.5 & 2.5  & 5  & 15 & 17.5 & 15 & 25 & 7.5  & 5 & 10.8 \\
        \hline
        Detection Fail & 0 & 0  & 0  & 2.5     & 0 & 0  & 0  & 0  & 0  & 5  & 0   & 2.5  & 10 & 1.5 \\
        \hline
    \end{tabular}
    \caption{Handover experiment outcomes, in percentage, categorised by the object. Each object was handed 40 times.}
    \label{table:eval}
    \vspace{-0.3cm}
\end{table*}

\subsection{Handover Experiments}
\label{sec:handover_experiments}

\begin{figure}[t!]
    \centering
    \includegraphics[trim={0cm 0.3cm 2cm 0.5cm},clip,width=1.0\columnwidth]{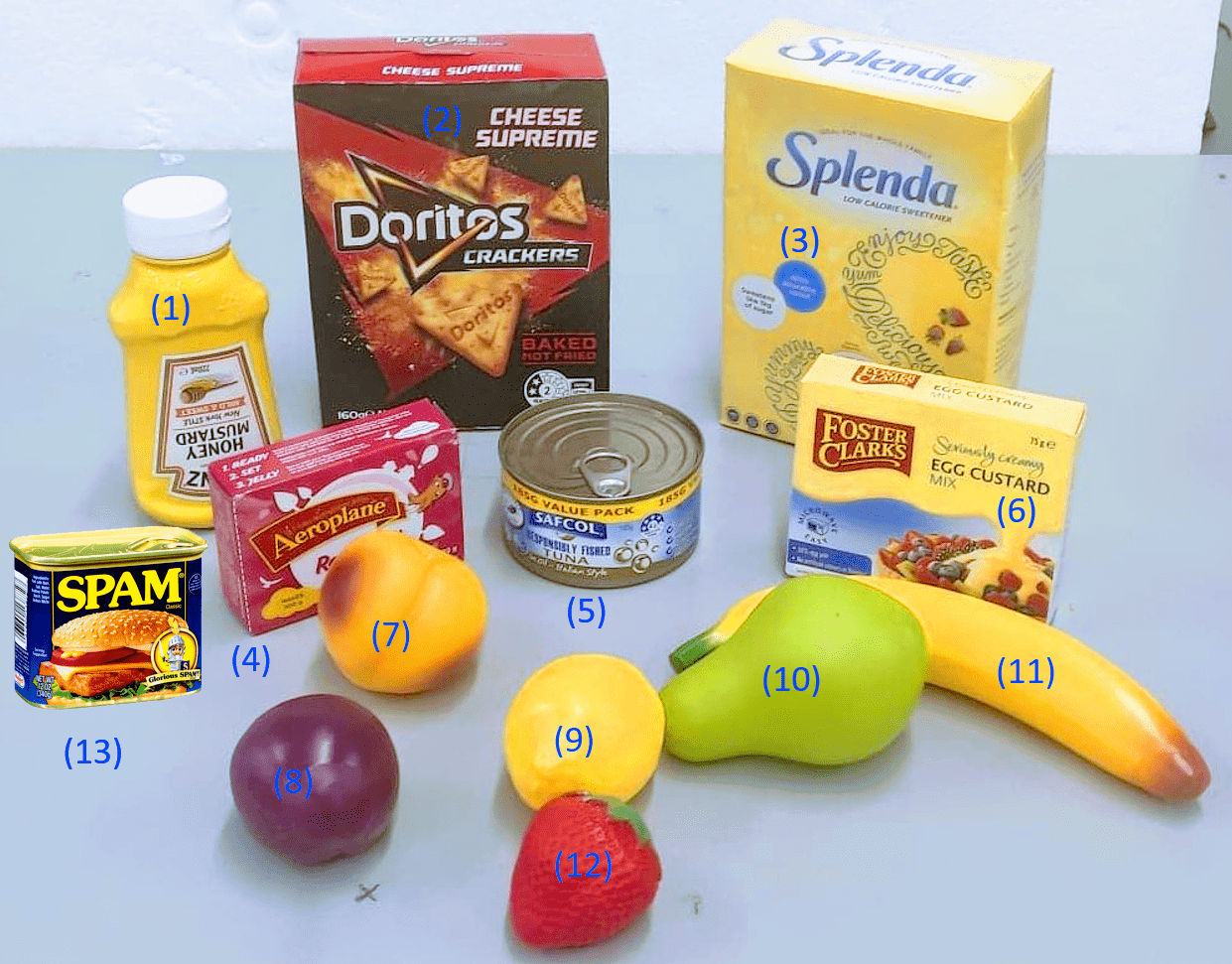}
    \caption{Test objects: (1) mustard, (2) biscuits, (3) sugar, (4) jelly, (5) tuna, (6) custard, (7) peach, (8) plum, (9) lemon, (10) pear, (11) banana, (12) strawberry, (13) spam. Fruit items were plastic.}
    \label{fig:eval_objects}
\end{figure}

The objective of the experiments is the successful transfer of an object from the human partner to the robot, while satisfying the design principles stated in Sec.~\ref{section_principles}. A set of 13 test objects was chosen (shown in Fig.~\ref{fig:eval_objects}), which consists of the food items in the YCB object set \cite{YCB2015} that fit to the Panda robot gripper. 

Four users, who are the developers of the system, were tasked to hand over the objects to the robot without instructions about how they should present the object. The users handed over each object 10 times consecutively before moving to the next one, which allowed users to adjust their handover behavior. The experiment consisted of a total of 520 handovers. An experiment observer labeled each trial with one of four outcomes:
\begin{itemize}
\setlength\itemsep{0pt}
    \item Success: The item was safely transferred to the robot.
    \item Safety Stop: The experiment observer stopped the robot if they thought that the robot might not act safely.
    \item Grasping Fail: The robot acted safely, but failed to receive the object.
    \item Detection Fail: The object is not detected within 30 sec.
\end{itemize}

\subsection{Quantitative Results}
\label{sec:evaluation_results}

As can be seen in Table~\ref{table:eval}, the overall success rate of all handovers was 81.9$\%$. The most common failure mode was the robot failing to grasp the object from the human (10.8$\%$ of the trials), which can be attributed to two causes. First, the selected grasp point was not suitable for grasping the object, \textit{e.g.} when the object at that point was larger than the gripper width. Secondly, the robot driver module automatically aborted the handover due to the human changing the hand's position during the handover and thereby getting too close to the grasp point, risking a collision with the robot.
In 5.7$\%$ of the trials the observer invoked the safety stop. This was most commonly due to the human moving the object after the handover had started. The experiment failed due to the object not being detected only 1.6$\%$ of the time. The results were fairly consistent among all objects, the spam and pear objects had the highest success rate with 92.5$\%$, whereas strawberry, biscuits and jelly objects performed the worst with 75$\%$ success rate. In its current configuration, the handover is not initiated until the object detection recognizes an object in the human's hand. While this is an intended safety feature, it causes problems if the object cannot be detected consistently. In our object set, the jelly object could not be detected consistently especially when it is oriented in a way where no distinguishing features are visible to the camera. Unfortunately, a comparative analysis of the evaluations results with the approaches in section \ref{section_related_work} is not feasible as the stated approaches have fundamentally different initial conditions (\textit{e.g.} only known objects, human responsible for ensuring safety, etc.). Therefore, a direct comparison is omitted.

\subsection{Qualitative Results and Future Work}
\label{sec:limit_internal_test}

\begin{figure}[bht!]
    \begin{subfigure}{0.32\columnwidth}
        \includegraphics[trim={2cm 0.5cm 1cm 1.5cm},clip,width=\textwidth]
            {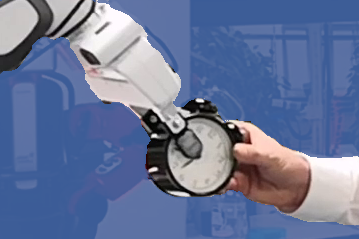}
        \label{fig:hold1}
    \end{subfigure}
    \vspace{-8pt}
    \begin{subfigure}{0.32\columnwidth} 
        \includegraphics[trim={1.7cm 1cm 1.3cm 1cm},clip,width=\textwidth]
            {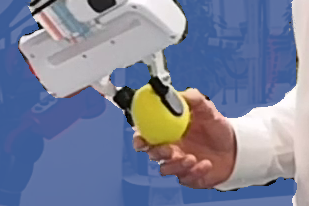}
        \label{fig:hold2}
    \end{subfigure}
    \begin{subfigure}{0.32\columnwidth} 
        \includegraphics[trim={2cm 1cm 1cm 1cm},clip,width=\textwidth]
            {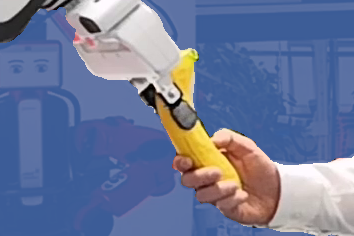}
        \label{fig:hold3}
    \end{subfigure}
    \begin{subfigure}{0.32\columnwidth} 
        \includegraphics[trim={1.7cm 0.5cm 1.3cm 1.5cm},clip,width=\textwidth]
            {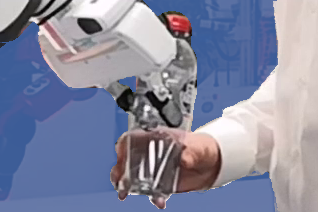} 
        \label{fig:hold4}
    \end{subfigure}
    \vspace{-8pt}
    \begin{subfigure}{0.32\columnwidth} 
        \includegraphics[trim={1.5cm 1cm 1.5cm 1cm},clip,width=\textwidth]
            {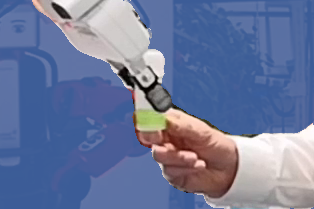} 
        \label{fig:hold5}
    \end{subfigure} 
    \begin{subfigure}{0.32\columnwidth} 
        \includegraphics[trim={2.5cm 1cm 0.5cm 1cm},clip,width=\textwidth]
            {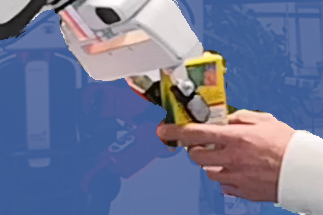} 
        \label{fig:hold6}
    \end{subfigure}
    \begin{subfigure}{0.32\columnwidth} 
        \includegraphics[trim={1.5cm 0.3cm 1.5cm 1.7cm},clip,width=\textwidth]
            {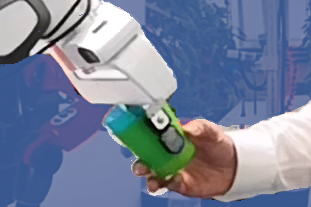} 
        \label{fig:hold7}
    \end{subfigure}
    \vspace{-8pt}
    \begin{subfigure}{0.32\columnwidth} 
        \includegraphics[trim={2cm 1.2cm 1cm 0.8cm},clip,width=\textwidth]
            {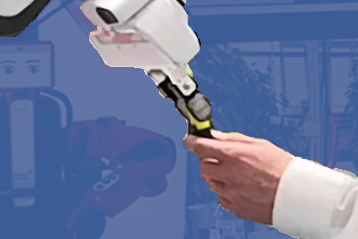} 
        \label{fig:hold8}
    \end{subfigure}
    \begin{subfigure}{0.32\columnwidth} 
        \includegraphics[trim={2.5cm 1cm 0.5cm 1cm},clip,width=\textwidth]
            {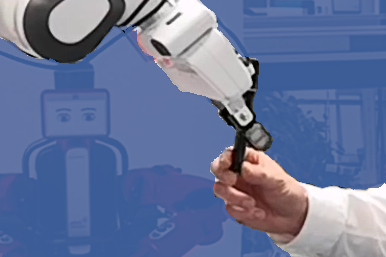} 
        \label{fig:hold9}
    \end{subfigure}
    \begin{subfigure}{0.32\columnwidth} 
        \includegraphics[trim={2.2cm 0.8cm 0.8cm 1.2cm},clip,width=\textwidth]
            {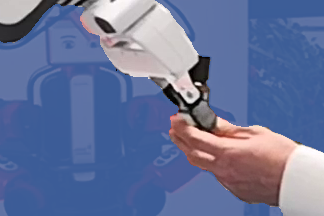} 
        \label{fig:hold10}
    \end{subfigure}
    \vspace{-8pt}
    \begin{subfigure}{0.32\columnwidth} 
        \includegraphics[trim={2cm 0.3cm 0cm 1cm},clip,width=\textwidth]
            {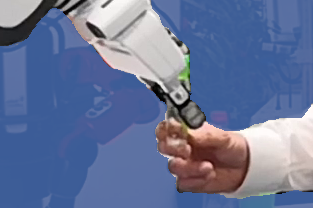} 
        \label{fig:hold11}
    \end{subfigure}
    \begin{subfigure}{0.32\columnwidth} 
        \includegraphics[trim={2cm 1cm 1cm 1cm},clip,width=\textwidth]
            {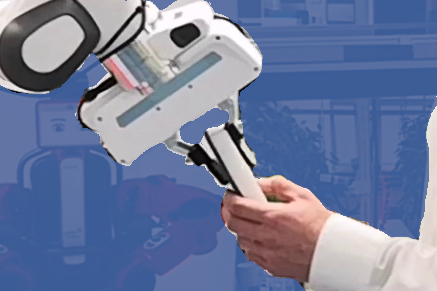} 
        \label{fig:hold12}
    \end{subfigure}
    \begin{subfigure}{0.32\columnwidth} 
        \includegraphics[trim={2cm 0.5cm 1cm 1.5cm},clip,width=\textwidth]
            {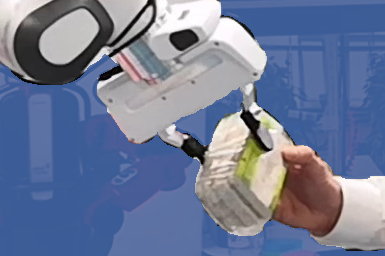} 
        \label{fig:hold13}
    \end{subfigure}
    \vspace{-8pt}
    \begin{subfigure}{0.32\columnwidth} 
        \includegraphics[trim={2cm 1cm 1cm 1cm},clip,width=\textwidth]
            {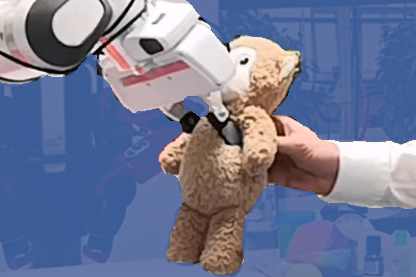} 
        \label{fig:hold14}
    \end{subfigure}
    \begin{subfigure}{0.32\columnwidth} 
        \includegraphics[trim={2cm 0.5cm 1cm 1.5cm},clip,width=\textwidth]
            {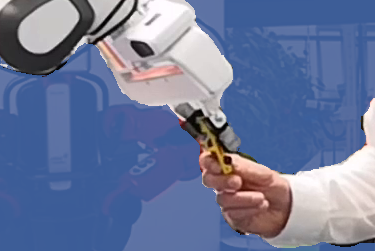} 
        \label{fig:hold15}
    \end{subfigure}
    \centering
    
    \caption{Handover examples of various objects as part of internal testing.}
    \label{fig:holds}
\end{figure}

Besides the reported pilot study, additional internal tests were conducted to identify weaknesses that should be addressed in future work. During these tests, ten robotics researchers not involved in the system's development handed the system a variety of everyday objects not belonging to any dataset used for training (see Fig.~\ref{fig:holds}).
Thereby, we varied (1) the user's position (e.g. in front of, beside, behind the robot, etc.), (2) the passed objects (e.g. mass, size, shape, transparency, etc.), (3) their presentation (e.g. with different orientations, with one and both hands, several objects at the same time, etc.), (4) the background behind the user (e.g. flat surface, lab environment, other people), and (5) the movement speed of the human hand (e.g. none, slow or fast). Compared to the pilot study, these internal tests did not have a strict test procedure but were intended to push the system's limits with respect to the five design principles (see Sec.~\ref{section_principles}).

Both the user study and the further tests revealed two limitations of our approach. 
Resolving these issues will lead to an increased success rate and a more natural feeling for the human partner. First, we believe that a closed-loop approach will significantly reduce the safety stop and grasping failures. A viable option would be an approach that tracks the handed object in 3D in real-time and constantly updates the location of the transfer point and the gripper's orientation.
Secondly, we found that the object detector is a bottleneck for our approach. We observed that sometimes the object is not detected, which prompts the user to move the object around in the hope that the robot would recognize it. Furthermore, the human partner may lose patience with the duration of the handover starting period. Both of these factors take away the naturalness of the human-robot interaction as well as reducing the willingness for further collaboration. 
To resolve this issue, an algorithm is needed that can determine if the user's hand is empty, if the user is holding an object without the intention to hand it over, or if the user is presenting an object to be handed over \cite{kwan2020gesture}. Further, this algorithms needs to be able to differentiate between the user's hand and the object as well as to determine the object's outline.

\addtolength{\textheight}{-6.6cm} 

\section{Conclusion} 
\label{section_conclusion}

We presented an approach for object-independent human-to-robot handovers using real-time robotic vision. The system's object-independent grasp selection allows the handover of previously unseen objects independently of their orientation, presentation, and background. Thanks to deep-learning based perception modules we can segment hands and human body parts, thus enabling safe handovers. The experiments, as well as internal tests have shown the system's robustness with respect to the requirements outlined in the design principles (Sec.~\ref{section_principles}). Nevertheless, it is not free from error and further development is required before it can be used in practice without supervision. We will do so by incorporating the modifications outlined in Sec.~\ref{sec:limit_internal_test}.


\bibliographystyle{IEEEtran}

\bibliography{bibliography} 


\end{document}